\documentclass[journal]{IEEEtran}
\usepackage{cite}
\usepackage{amsmath,amssymb,amsfonts}
\usepackage{algorithm}
\usepackage{algpseudocode}
\usepackage{graphicx}
\usepackage{hyperref}
\hypersetup{hidelinks=true}
\usepackage{textcomp}
\usepackage{xcolor}
\usepackage{mathptmx}
\usepackage{multirow}
\usepackage{booktabs}
\usepackage[justification=centering]{caption}
\usepackage{makecell}
\usepackage{multicol}
\usepackage{setspace}
\usepackage{lipsum} 
\usepackage{subfigure}
\usepackage{tabularx}
\usepackage{float}
\usepackage{bm}

\ifCLASSINFOpdf

\begin{document}
%

\title{Linguistic Fuzzy Information Evolution with Random Leader Election Mechanism \\for Decision-Making Systems}

\author{Qianlei Jia, Witold Pedrycz, \emph{Life Fellow, IEEE}
\thanks{Corresponding author (Qianlei Jia)}
\thanks{Qianlei Jia is with the School of Mechanical and Aerospace Engineering, Nanyang Technological University, 639798, Singapore, (e-mail: jiaql@mail.nwpu.edu.cn)}
\thanks{Witold Pedrycz is with the Department of Electrical and Computer Engineering, University of Alberta, Edmonton T6R 2V4 AB, Canada, the
School of Electro-Mechanical Engineering, and also the Faculty of Engineering, King Abdulaziz University, Jeddah 21589, Saudi Arabia (e-mail:
wpedrycz@ualberta.ca).}
}

\maketitle

\begin{abstract}
\par Linguistic fuzzy information evolution is crucial in understanding information exchange among agents. However, different agent weights may lead to different convergence results in the classic DeGroot model. Similarly, in the Hegselmann-Krause bounded confidence model (HK model), changing the confidence threshold values of agents can lead to differences in the final results. To address these limitations, this paper proposes three new models of linguistic fuzzy information dynamics: the per-round random leader election mechanism-based DeGroot model (PRRLEM-DeGroot), the PRRLEM-based homogeneous HK model (PRRLEM-HOHK), and the PRRLEM-based heterogeneous HK model (PRRLEM-HEHK). In these models, after each round of fuzzy information updates, an agent is randomly selected to act as a temporary leader with more significant influence, with the leadership structure being reset after each update. This strategy increases the information sharing and enhances decision-making by integrating multiple agents' evaluation information, which is also in line with real life (\emph{Leader is not unchanged}). The Monte Carlo method is then employed to simulate the behavior of complex systems through repeated random tests, obtaining confidence intervals for different fuzzy information. Subsequently, an improved golden rule representative value (GRRV) in fuzzy theory is proposed to rank these confidence intervals. Simulation examples and a real-world scenario about space situational awareness validate the effectiveness of the proposed models. Comparative analysis with the other models demonstrate our ability to address the echo chamber and improve the robustness.

\end{abstract}

\begin{IEEEkeywords}
Linguistic Fuzzy Information, Per-Round Random Leader Election Mechanism, Decision-Making, Golden Rule Representative Value (GRRV).
\end{IEEEkeywords}

\IEEEpeerreviewmaketitle
\renewcommand\arraystretch{1.2}

\section{Introduction}
\IEEEPARstart{F}{uzzy} information opinion dynamics can help us predict and explain the evolution of social phenomena, such as the formation of public opinion, policy-making, and market fluctuations \cite{10654535}. The related research provides important insights for the development of effective policies and management strategies \cite{9806378, 9780145}.

\par In general, the fuzzy opinion dynamics models are mainly divided into
binary models \cite{muslim2024mass} and continuous models \cite{liang2016fusion}. Binary models mainly include the voter model \cite{jiao2021active}, the Sznajd model  \cite{benatti2020opinion}, and the majority-rule model \cite{forgerini2023directed}. These models are advantageous due to their simplicity, intuitiveness, and applicability to large-scale group simulations and analyses. However, they also have limitations, as overly simplified models may fail to accurately capture variations in evaluation information within a continuous range and overlook complex interactions and information exchange among agents. Continuous models mainly include the DeGroot model and the bounded confidence model. The DeGroot model is based on the principle of linear weighted averaging, assuming that agents gradually adjust their evaluation information based on the consistency with the information of their neighbors until consensus is reached \cite{9028187}. The advantage of this model lies in its simplicity and intuitiveness, making it easy to understand and implement. However, it assumes that all agents can accept and adopt the opinions of others, overlooking the complex interactions between agents. The bounded confidence model introduces the concept of ``bounded confidence," where agents only interact with neighbors holding similar opinions and update their own evaluation information only when their confidence exceeds a certain threshold \cite{jiang2018predicting}. The HK model is a typical representative of the bounded confidence model, which considers differences in confidence between agents, making it closer to real-world scenarios and possessing good mathematical properties \cite{su2021noise}. However, the bounded confidence model also has some drawbacks, such as sensitivity to parameter settings, and may fail to capture complex dynamics of fuzzy information change in certain situations \cite{9233959}. Additionally, the majority of opinion dynamics models are applicable in numerical environment. However, in the real-world situations, linguistic fuzzy expressions, such as ``\emph{good}" and ``\emph{poor}", often align better with our natural way of communication \cite{10526446, 10495183}. Considering the above, detailed research on linguistic opinion dynamics models is necessary.

\par In research related to information evolution, leadership has attracted attention. The current research about leadership is divided into three main parts. First is the analysis of the impact of leaders on the evolution of the opinion \cite{LIU201534}. The second part is the consideration of different methods to model social agents \cite{almeida2019optimal}. Another part involves combining the leader-follower architecture with different types of opinion dynamics models \cite{dabarera2016consensus, 9711916}. These methods based on the leadership mechanism provide a new perspective for studying uncertain information evolution. However, the traditional opinion dynamics models considering the leadership mechanism face three main limitations: (1) Usually, studies predetermine the identity of leaders based on specific criteria. Typically, individuals with higher centrality in social networks or those with significant personal prestige and reputation are identified as leaders. However, this approach overly simplifies leadership, which should be context-dependent and not solely determined by static indicators such as centrality or reputation. (2) Many studies assume that the leaders remain unchanged throughout the opinion update process. This approach simplifies the construction and analysis of models but overlooks a vital reality: in actual systems, the identity and influence of leaders are dynamic. For example, leadership within an organization can change due to individual abilities, support from team members, or changes in the external environment. The static setting of leaders leads to models that cannot accurately reflect the actual dynamics of the system, especially when dealing with complex systems or rapidly changing environments. (3) If the leaders remain unchanged or only change slightly, it will affect the model's robustness and significantly reduce the accuracy of the final results. When leaders provide the linguistic fuzzy information, it is usually based on their subjective knowledge and judgment. However, leaders can inevitably hold subjective biases or unintentionally provide incorrect information, significantly increasing the risk of subjectivity in the information. Due to the traditional mechanism for setting up leaders, this risk is difficult to avoid.

\par Motivated by the challenges mentioned above, this paper aims to propose three new linguistic fuzzy information evolution models based on the per-round random leader election mechanism (PRRLEM). The idea comes from a recent study published in \emph{Nature Physics} \cite{gomez2022intermittent}, which revealed that sheep can exhibit collective wisdom through random, alternating leadership behavior. After each grazing session, a sheep is randomly chosen as the leader, and others follow in succession, with this leadership structure resetting after each feeding cycle. The advantages of this mechanism are the ability to provide efficient collective decision-making and adaptive advantages for the system. The system can flexibly adjust leaders in different situations through random, alternating leadership behavior, thus avoiding over-reliance on a single leader and reducing the impact of leader single-point failures on the system. Additionally, this mechanism promotes diversity and sharing of fuzzy information within the system, as each collective movement integrates opinions and experiences from different agents. Based on this mechanism, this paper proposes three new models: the PRRLEM-DeGroot model (PRRLEM-DeGroot), the PRRLEM-homogeneous HK model (PRRLEM-HOHK), and the PRRLEM-heterogeneous HK model (PRRLEM-HEHK). Moreover, considering that the results of some models will be significantly different with different parameter settings, this paper employs the Monte Carlo method and confidence intervals to address the randomness and uncertainty inherent in fuzzy information. Finally, a new golden rule representative value (GRRV) is proposed to rank the opinions. To our knowledge, no scholars have employed PRRLEM, Monte Carlo method, and GRRV to study the linguistic fuzzy information evolution. Simulation examples and a real-life case have validated the effectiveness of the three algorithms proposed in this paper. A detailed comparison further confirms that the proposed algorithms have higher accuracy and robustness and can also effectively address the issue of echo chambers in information dissemination.

\par The structure of the paper is as follows: In Section \uppercase\expandafter{\romannumeral2}, the three new fuzzy information evolution models are proposed. In Section \uppercase\expandafter{\romannumeral3}, a series of examples are presented. In Section \uppercase\expandafter{\romannumeral4}, a detailed comparison analysis is conducted.   The conclusion is given in Section \uppercase\expandafter{\romannumeral5}.

\section{Models and Definitions}
\par In this section, we propose the PRRLEM-DeGroot, PRRLEM-HOHK, and PRRLEM-HEHK, complete with detailed operational steps and pseudocode for each.
\subsection{PRRLEM-DeGroot}
\par The traditional DeGroot model shows certain limitations in simulating opinion formation, primarily due to the uniqueness of the final results. To address this issue, we introduce a random leadership mechanism \cite{gomez2022intermittent} to enhance decision-making diversity. The algorithm steps are as follows:
\par \emph{Step 1: Initialization}
\par In daily life, using linguistic terms such as ``\emph{good}" and ``\emph{bad}" for expression more closely aligns with our living habits. Therefore, scholars proposed the concept of linguistic term set (LTS). It is a structured collection of linguistic terms, essentially descriptive labels, used to express judgments, opinions, or values in a way that mimics natural language. An LTS typically consists of a finite and ordered set of terms $H=\lbrace h_\xi \vert \xi=0,1,\cdots,2\Phi,\vert \Phi\in \mathbb{N}^* \rbrace$ \cite{9204382} (Typically, the number of linguistic terms is odd). For two linguistic terms $h_i$ and $h_j$, it should satisfy the four constraints: (1) $h_i \leq h_j$ if and only if $i \leq j$; The negation operation $neg(h_i)=h_j$ if $i + j=2\Phi$; (3) If $i \geq j$, then $\max \lbrace h_i,h_j \rbrace=h_i$; (4) If $i \geq j$, then $\min \lbrace h_i,h_j \rbrace=h_j$. In this study, the total number of agents is $N$, with each agent $e_i$ having an initial linguistic opinion $\kappa_i(t)$ at time $t=1$. The number of Monte Carlo tests is $M$, and the iteration number in opinion dynamics is $T$.

\par \emph{Step 2: Opinion Conversion}
\par After obtaining linguistic fuzzy information, we first need to convert this information into corresponding specific numerical values to facilitate further information updates and analysis. The following transformation is widely used \cite{JIA2023105610}.
\begin{equation} 
\label{eq:trans}
\theta_\xi=\left\{ \begin{array}{rcl} \frac{a^\Phi-a^{\Phi-\xi}}{2a^\Phi-2}, & & {0 \leq \xi \leq \Phi}\\ 
\frac{a^\Phi+a^{\xi-\Phi}-2}{2a^\Phi-2}, & & {\Phi <\xi\leq 2\Phi } \end{array} 
\right. \end{equation}
\par where $\theta_\xi \in [0,\,1]$. The maximum value of $\theta_\xi$ is 1, obtained when $\xi$ is $2\Phi$, and the minimum value is 0, obtained when $\xi$ is 0.

\par \emph{Step 3: Weight Allocation}
\par As previously analyzed, to ensure collective wisdom and avoid over-reliance on a single leader, we randomly select an agent $j$ as the leader and assign it a random weight $w_j \in [0,1]$ when these agents update their information. 
\begin{equation}
\label{eq:Sheep_opinion}
\omega_j \sim \text{Uniform}(0,1)
\end{equation}
\par The weight of the remaining agents is:
\begin{equation}
\label{eq:theta}
\omega_k=\frac{1 - \omega_j}{N - 1}
\end{equation}
\par Apart from the randomly selected leader, the weights of the remaining followers are the same in the opinion update.

\par \emph{Step 4: Opinion Update}
\par In the DeGroot model, the agents do not have bounded confidence and trust all the agents. According to \cite{zhou2020two, wu2022mixed}, the numerical value of $e_i$'s opinion at time $t+1$ is:
\begin{equation}
\label{eq:update}
y_i(t+1)=\omega_j y_j(t)+\sum_{k=1,k\neq j}^N \omega_k y_k(t)
\end{equation}
\par Actually, when $y_i(t+1)\geq 1$, $y_i(t+1)$ exceeds the domain range, and $y_i(t+1)\geq 1$ takes the maximum value of the entire domain, which is $h_{2\Phi}$. Similarly, when $y_i(t+1)\leq 0$, it also means $y_i(t+1)$ exceeds the domain range, and therefore $y_i(t+1)$ takes the minimum value of the entire domain, which is $h_0$. When $y_i(t+1)$ is within the domain, we need to calculate the distance between $y_i(t+1)$ and each linguistic term in $H=\lbrace h_\xi \vert \xi=0,1,\cdots,2\Phi,\vert \Phi\in \mathbb{N}^* \rbrace$, and the term closest to $y_i(t+1)$ will be recorded as the linguistic opinion $\kappa_i(t+1)$ after opinion update. The linguistic expression of $y_i(t+1)$ can be obtained:
\begin{equation}
\label{eq:linguistic}
\begin{aligned}
\kappa_i(t\!+\!1)\!=\!\left\{ \begin{array}{lr} h_{2\Phi},\quad\quad\quad y_i(t+1) \geq 1;\\ 
\underset{\theta_\xi \in H}{\text{argmin}}\, \vert y_i(t+1)-\theta_\xi \vert,\;\, 0<y_i(t+1) <1;\\ 
h_0,\qquad\;\quad\; y_i(t+1) \leq 0. \end{array} 
\right.
\end{aligned}
\end{equation}

\par \emph{Step 5: Iterative Simulation and Result Recording}
\par Repeat the process from \emph{Step 2} to \emph{Step 4} within the framework of opinion dynamics. In each iteration, we simulate the interactions of $N$ agents to model the opinion evolution, continuing until the number of iterations $T$ is reached, completing a round of the Monte Carlo test. At the end of this round, we record the final opinions of the $N$ agents as $\kappa_i(T)$. To ensure the robustness and reliability of the results, repeat the Monte Carlo test $M$ times, each time independently simulating the evolution of agents' opinions. We then record the linguistic terms $h_\xi$ that occur in the entire Monte Carlo tests and then count the number of times, denoted as  $Q_\xi$.

\par \emph{Step 6: Monte Carlo Analysis}
\par After obtaining the number of times of each opinion, the next step involves conducting a mathematical analysis to determine the final results. Therefore, we calculate each opinion's proportion $p_\xi = \frac{Q_\xi}{MN}$. Choose a confidence level (commonly used is 95\%) corresponding to a confidence coefficient $Z$ (for a 95\% confidence level, $Z \approx 1.96$) \cite{hazra2017using}. The confidence interval is:
\begin{equation}
\label{eq:Monte}
\begin{aligned}
CI_\xi = p_\xi \pm Z \cdot \sqrt{\frac{p_\xi(1 - p_\xi)}{MN}}
\end{aligned}
\end{equation} 
\par The confidence interval provides an interval estimate for each opinion. The width of the confidence interval reflects the uncertainty in the estimated probability of an event occurring, with a narrower confidence interval indicating higher estimation precision, while a wider confidence interval suggests more significant uncertainty. At the same time, a higher midpoint of the confidence interval indicates a higher estimated probability of the corresponding event occurring.
\par \emph{Step 7: Confidence Intervals Ranking and Results Determination}
\par After obtaining the confidence intervals, the next step involves comparing different values to determine the most likely opinions of each agent. This task essentially compares and ranks different confidence intervals. Researchers have employed various methods to address this issue, including sorting based on the midpoint of confidence intervals, the width of confidence intervals, and the degree of overlap between confidence intervals. However, these methods have limitations. For instance, sorting solely based on the midpoint of confidence intervals may overlook the estimate's uncertainty, while sorting based on the width of confidence intervals may neglect the actual size of the estimate. To overcome these limitations, we employ the golden rule representative value (GRRV) proposed by Yager \cite{7073630}. 

\par Before that, it is necessary to introduce the GRRV. The principle of GRRV is to assign a representative scalar value, recorded as $Rep$, to each interval $x_i = [a_i, b_i],\,a_i,b_i\in [0,1]$ \cite{7934318, 7073630}. Usually, $x_i$ is considered to be preferred over $x_j$ when $ Rep (x_i)> Rep (x_j) $. To obtain the representative value, the following rules are defined \cite{7073630}.
\par \emph{Rule 1}: IF $m_i$ is \emph{large} and $r_i$ is \emph{small}, THEN $Rep(x_i) = 1$;
\par \emph{Rule 2}: IF $m_i$ is \emph{large} and $r_i$ is \emph{large}, THEN $Rep(x_i) = 1/2$;
\par \emph{Rule 3}: IF $m_i$ is \emph{small} and $r_i$ is \emph{large}, THEN $Rep(x_i) = 1/2$;
\par \emph{Rule 4}: IF $m_i$ is \emph{small} and $r_i$ is \emph{small}, THEN $Rep(x_i) = 0$.
\par where $m_i=(a_i+b_i)/2$ and $r_i=b_i-a_i$ indicate the mean and range of these intervals. \emph{large} and \emph{small} are linear fuzzy sets $L$ and $S$. $L(y)=y$ and $S(z)=1-z$ are set in \cite{7073630}. 
\par To derive the representative value, the TSK fuzzy model is employed. For a TSK fuzzy system with $d$ inputs, 1 output, and $K$ rules, the fuzzy rules are \cite{1988Structuridentificatione, 10480235}:
\par \emph{Rule $k$}: IF $q_1$ is $A_1^k$ $\wedge$ $q_2$ is $A_2^k$ $\wedge$ $\cdots$ $\wedge$ $q_d$ is $A_d^k$;
\par $\quad\quad\;\;\;\;$ THEN $y_k(Q)=p_0^k+p_1^k q_1+\cdots+p_d^k q_d$.
\par where $Q=\lbrace q_1, q_2,\cdots,q_d \rbrace$ indicates the predecessor variables. $A_d^k$ represents the fuzzy set of $q_d$ in $Rule\;\, k$. $y_k(Q)$ is the post-component variable. When the input is $Q=\lbrace q_1, q_2,\cdots,q_d \rbrace$, the output is:
\begin{equation}
\label{TSK}
y=\frac{\sum_{i=1}^K \prod_{j=1}^d A_j^i(q_j)y_i}{\sum_{i=1}^K \prod_{j=1}^d A_j^i(q_j)}
\end{equation}
\par Based on (\ref{TSK}), the representative value of $x_i=[a_i,b_i]$ can be derived.
\begin{equation}
\begin{aligned}
Rep(x_i)=\!\!\frac{m_i(1-r_i)+\frac{1}{2}m_i r_i+\frac{1}{2}(1-m_i)r_i+0(1-m_i)(1-r_i)}{m_i(1-r_i)+m_i r_i+(1-m_i)r_i+(1-m_i)(1-r_i)} \\
=m_i+(\frac{1}{2}-m_i)r_i \qquad\qquad\qquad\qquad\qquad\;\qquad\quad\quad\,\,
\\=\frac{1}{2}(a_i^2+2b_i-b_i^2)\qquad\qquad\qquad\qquad\qquad\;\;\,\,\;\qquad\quad
\end{aligned}
\end{equation}
\par Similarity, the representative value $Rep(CI_\xi)$ of the confidence interval $CI_\xi$ is:
\begin{equation}
\label{eq:GRRV}
\begin{aligned}
Rep(CI_\xi) =\frac{\min{CI_\xi}+\max{CI_\xi}}{2} \quad\quad\quad\quad\quad\quad\quad\quad\quad\\+ \Big(\frac{1}{2}-\frac{\min{CI_\xi}+\max{CI_\xi}}{2}\Big)(\max{CI_\xi}-\min{CI_\xi}) 
\end{aligned}
\end{equation} 
\par After obtaining the representative value for each interval, ranking the different confidence intervals becomes straightforward. A confidence interval with a more considerable representative value is considered to have a higher probability of occurrence. Unlike other methods, GRRV considers both the midpoint and the width. We select the highest value $Rep_{\max}(CI_\xi)$ as the final opinion of the system. In this way, we ensure that each opinion can be considered. Furthermore, by repeating this process in multiple tests, we fully consider the inherent randomness of the system, ensuring the robustness of the algorithm's results. The pseudocode and flowchart of the algorithm are shown in \textbf{Algorithm 1}.

\begin{algorithm}
\caption{PRRLEM-DeGroot and Monte Carlo Method}\label{alg:opinion_dynamics}
\begin{algorithmic}[1]
\State \textbf{Input:} $N, M, T, \Phi, a$
\State \textbf{Output:} The most likely opinions for each agent
\State \emph{Step 1: Initialization}
\State Define a linguistic term set $H = \lbrace h_\xi \vert \xi = 0, 1, \ldots, 2\Phi \rbrace$, where $\Phi \in \mathbb{N}^*$.
\State Initialize agent $e_i$ with a linguistic opinion $\kappa_i(t)$ at $t=1$.
\State Set the number of Monte Carlo simulations $M$ and the iteration limit $T$.
\State \emph{Step 2: Opinion Conversion}
\For{each agent $i$ from 1 to $N$}
    \State Convert linguistic opinions $\kappa_i(t)$ to $\theta_i(t)$.
\EndFor
\State \emph{Step 3: Weight Allocation}
\For{each iteration $t$ from 1 to $T$}
    \State Select an agent $j$ randomly as the influential agent.
    \State Assign random weight $w_j \in [0,1]$ to the selected agent.
    \State Assign weights $\omega_k = \frac{1 - w_j}{N - 1}$ to other agents.
\EndFor
\State \emph{Step 4: Opinion Update}
\For{each agent $i$ from 1 to $N$}
    \State Update agent $i$'s opinion based on the weighted sum of its own and the neighbors' opinions.
\EndFor
\State \emph{Step 5: Iterative Simulation and Result Recording}
\For{$m = 1$ to $M$}
    \For{$t = 1$ to $T$}
        \State Repeat \emph{Step 2} to \emph{Step 4}.
    \EndFor
    \State Record the final opinions of the $N$ agents, marking them as $\kappa_i(T)$.
\EndFor
\State \emph{Step 6: Monte Carlo Analysis}
\For{each linguistic term $h_\xi$}
    \State Tally occurrences in all Monte Carlo simulations, denoted as $Q_\xi$.
\EndFor
\State Calculate the proportion $p_\xi$ for each opinion.
\State \emph{Step 7: Confidence Intervals Ranking and Results Determination}
\For{each opinion $h_\xi$}
    \State Calculate the confidence interval $CI_\xi$.
    \State Apply the GRRV method to rank the confidence intervals.
\EndFor
\State Determine the most likely opinion for each agent.
\State \textbf{return} The ranked list of the most likely opinions for each agent
\end{algorithmic}
\end{algorithm}

\subsection{PRRLEM-HOHK}
\par Unlike the DeGroot model, agents only trust people whose opinions
are close to their own in the HK model. If all agents hold the same
confidence threshold $\varepsilon$, the HK model is homogeneous; otherwise, it is heterogeneous. Here, two novel HK models are proposed. The calculation process of PRRLEM-HOHK is shown below.

\par \emph{Step 1: Initialization}
\par Identical to PRRLEM-DeGroot, a finite and ordered LTS is defined. Each agent is initialized with a linguistic opinion.
\par \emph{Step 2: Opinion Conversion}
\par This step remains unchanged, focusing on converting linguistic opinions into numerical values.
\par \emph{Step 3: Confidence Set Determination}
\par The characteristic of HK model is that each agent has the confidence set. According to \cite{su2021noise}, the confidence set $I(e_i,Y(t))$ of $e_i$ at time $t$ is:
\begin{equation}
\label{eq:confidence_set}
I(e_i,Y(t))=\lbrace e_j\vert \vert y_i(t)-y_j(t)\vert\leq \varepsilon,e_j\in E\rbrace
\end{equation}
\par $\# I(e_i,Y(t))$ indicates the cardinality of $I(e_i,Y(t))$.

\par \emph{Step 4: Weight Allocation}
\par In the confidence set of $e_i$, one agent $e_j$ is randomly selected as the leader. The weight of $e_j$ is defined as:
\begin{equation}
\label{eq:Sheep_HKopinion}
\omega_j \sim \text{Uniform}(0,1)
\end{equation}
\par For the other agents in $I(e_i,Y(t))$, the weights are:
\begin{equation}
\label{eq:theta1}
\omega_k=\frac{1 - \omega_j}{\# I(e_i,Y(t))-1}
\end{equation}
\par Specifically, when $\varepsilon$ is very low, resulting in $I(e_i,Y(t))$ only including $e_i$. The weight $\omega_i$ during the opinion update is 1, reflecting the agent's complete trust in its own opinion without external trusted agents. Moreover, if the confidence sets of two agents are the same, i.e., $I(e_i, Y(t))=I(e_j, Y(t))$, then the weights of the agents in the confidence set are consistent when $e_i$ and $e_j$ update the opinions.

\par \emph{Step 5: Opinion Update}
\par In this step, agents update their opinions by considering only the opinions within their confidence set. The calculation formula is the same as PRRLEM-DeGroot.

\par \emph{Step 6: Iterative Simulation and Result Recording}
\par Repeat the process from \emph{Step 2} to \emph{Step 5} within the framework of opinion dynamics. It is worth emphasizing that compared to the DeGroot model, the agents in the HK model may not ultimately reach a consensus due to the confidence set. The leader selected for each round of opinion updates is random, and different rounds may result in different opinion groups, which is different from the DeGroot model in that all agents will certainly achieve consensus. For example, after one round of Monte Carlo, $e_1$, $e_2$, and $e_4$ may end up with the same opinion, but in the next round, only $e_1$ and $e_2$ have the same opinion. Therefore, we tally the linguistic terms $h_\xi$ that occur for $e_i$ throughout the entire Monte Carlo test and calculate the number of times each linguistic term occurs, denoted as $Q_\xi^{e_i}$.

\par \emph{Step 7: Monte Carlo Analysis}
\par For each agent, calculate the proportion of each opinion $p_\xi^{e_i} = \frac{Q_\xi^{e_i}}{M}$. Choose a confidence level (95\%). For each opinion, calculate the confidence interval using the formula:
\begin{equation}
\label{eq:Monte_HK}
\begin{aligned}
CI_\xi^{e_i} = p_\xi^{e_i} \pm Z \cdot \sqrt{\frac{p_\xi^{e_i}(1 - p_\xi^{e_i})}{M}}
\end{aligned}
\end{equation} 
\par \emph{Step 8: Confidence Intervals Ranking and Results Determination}
\par The representative value  $Rep(CI_\xi^{e_i})$ of the confidence interval $CI_\xi^{e_i}$ is:
\begin{equation}
\label{eq:GRRV1}
\begin{aligned}
Rep(CI_\xi^{e_i}) =\frac{\min{CI_\xi^{e_i}}+\max{CI_\xi^{e_i}}}{2} \quad\quad\quad\quad\quad\quad\quad\quad\quad\\+ \Big(\frac{1}{2}-\frac{\min{CI_\xi^{e_i}}+\max{CI_\xi^{e_i}}}{2}\Big)(\max{CI_\xi^{e_i}}-\min{CI_\xi^{e_i}}) 
\end{aligned}
\end{equation} 
\par The highest value $Rep_{\max}(CI_\xi^{e_i})$ is determined as the final opinion of the agent $e_i$. The pseudocode and flowchart of the algorithm are shown in \textbf{Algorithm 2}.
\begin{algorithm}
\caption{PRRLEM-HOHK and Monte Carlo Method}\label{alg:enhanced_opinion_dynamics}
\begin{algorithmic}[1]
\State \textbf{Input:} $N, M, T, \Phi, a$
\State \textbf{Output:} The most likely opinions for each agent

\State \emph{Initialization, Opinion Conversion:} Identical to \emph{Steps 1-2} in \textbf{Algorithm \ref{alg:opinion_dynamics}}

\State \emph{Step 3: Confidence Set Determination}
\For{each agent $i$ from 1 to $N$}
    \State Determine confidence set $I(e_i,Y(t))$ using threshold $\varepsilon$
\EndFor

\State \emph{Step 4: Weight Allocation}
\For{each agent $i$ from 1 to $N$}
    \State Randomly select a leader within the confidence set.
    \State Assign weights according to (\ref{eq:Sheep_HKopinion}) and (\ref{eq:theta1})
\EndFor

\State \emph{Step 5: Opinion Update:} Proceed as in \emph{Step 4} in \textbf{Algorithm \ref{alg:opinion_dynamics}}, but using the confidence set.

\State \emph{Step 6: Iterative Simulation, Result Recording} Same as \emph{Steps 5-6} in \textbf{\textbf{Algorithm \ref{alg:opinion_dynamics}}}, but incorporating the confidence set for each agent.

\State \emph{Step 7: Monte Carlo Analysis}
\For{each linguistic term $h_\xi$}
    \State Tally occurrences in all Monte Carlo simulations, denoted as $Q_\xi^{e_i}$.
\EndFor
\State Calculate the proportion $p_\xi^{e_i}$ for each opinion.

\State \emph{Step 8: Confidence Intervals Ranking and Results Determination}
\For{each agent $i$ from 1 to $N$}
    \State Apply GRRV method to rank the confidence intervals $CI_\xi^{e_i}$ using (\ref{eq:GRRV})
    \State Determine the final opinion for each agent based on the highest GRRV score.
\EndFor

\State \textbf{return} The ranked list of the most likely opinions for each agent
\end{algorithmic}
\end{algorithm}
\subsection{PRRLEM-HEHK}
\par Different agents may have varying confidence thresholds in the real world based on their individual characteristics and environmental backgrounds. PRRLEM-HOHK assumes that all agents have the same threshold $\varepsilon$. To better reflect reality, we improve it by allowing each agent to possess its confidence threshold $\varepsilon_i$ and form the PRRLEM-HEHK model. The pseudocode is shown in \textbf{Algorithm 3}. 
\begin{algorithm}
\caption{PRRLEM-HEHK and Monte Carlo Method}\label{alg:improved_enhanced_opinion_dynamics}
\begin{algorithmic}[1]
\State \textbf{Input:} $N, M, T, \Phi, a, \{\varepsilon_i\}_{i=1}^{N}$
\State \textbf{Output:} The most likely opinions for each agent

\State \emph{Initialization, Opinion Conversion, and Other Steps:} Refer to \textbf{Algorithm \ref{alg:enhanced_opinion_dynamics}}

\State \emph{Step 3: Individual Confidence Set Determination}
\For{each agent $i$ from 1 to $N$}
    \State Update to consider individual threshold $\varepsilon_i$:
    \State $I(e_i,Y(t))=\lbrace e_j\vert \vert y_i(t)-y_j(t)\vert\leq \varepsilon_i,e_j\in E\rbrace$
\EndFor

\State \emph{Further Steps:} Proceed as in Algorithm \ref{alg:enhanced_opinion_dynamics}, incorporating the unique $\varepsilon_i$ for each agent in the confidence set determination.

\State \textbf{return} The ranked list of the most likely opinions for each agent, accounting for individual trust thresholds.
\end{algorithmic}
\end{algorithm}
\begin{figure}[h]
\centering
\includegraphics[width=3.4in]{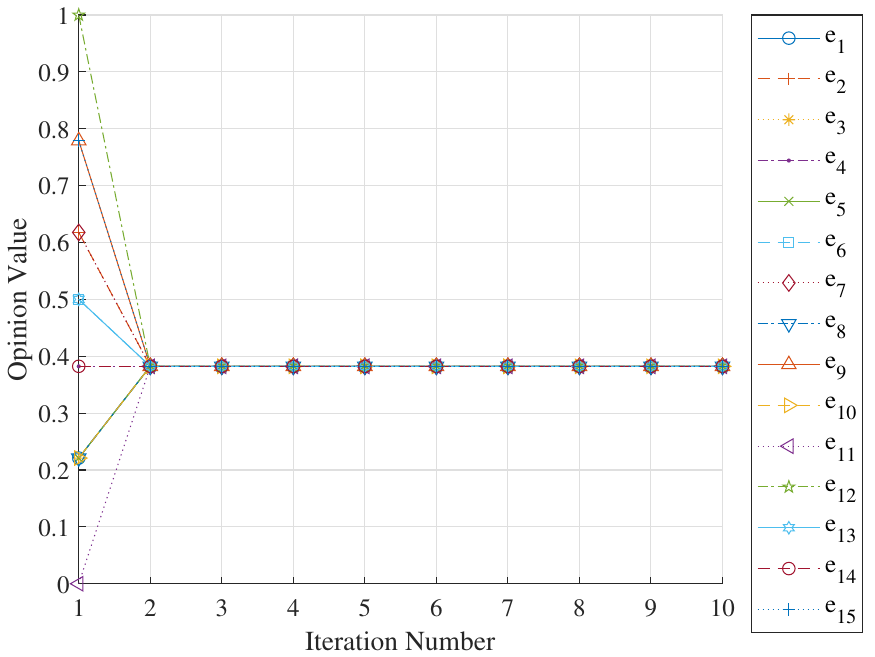}
\caption{Opinion Evolution of All Agents.}
\label{updateopi}
\end{figure}
\begin{figure}[h]
\centering
\includegraphics[width=3.4in]{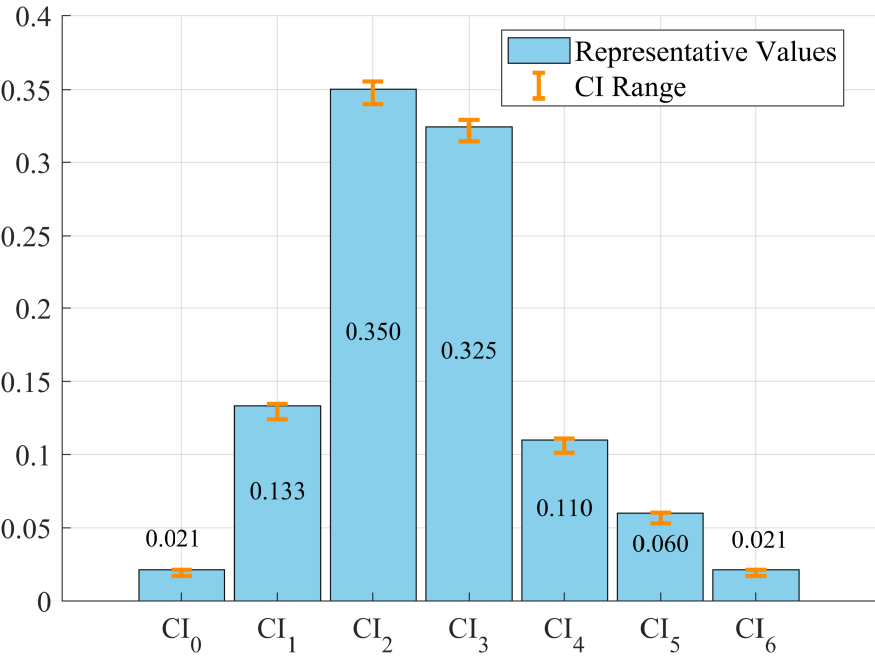}
\caption{Confidence Intervals and Representative Values of Each Opinion after Monte Varlo Tests.}
\label{proposed_DG}
\end{figure}
\begin{figure}[h]
\centering
\includegraphics[width=3.4in]{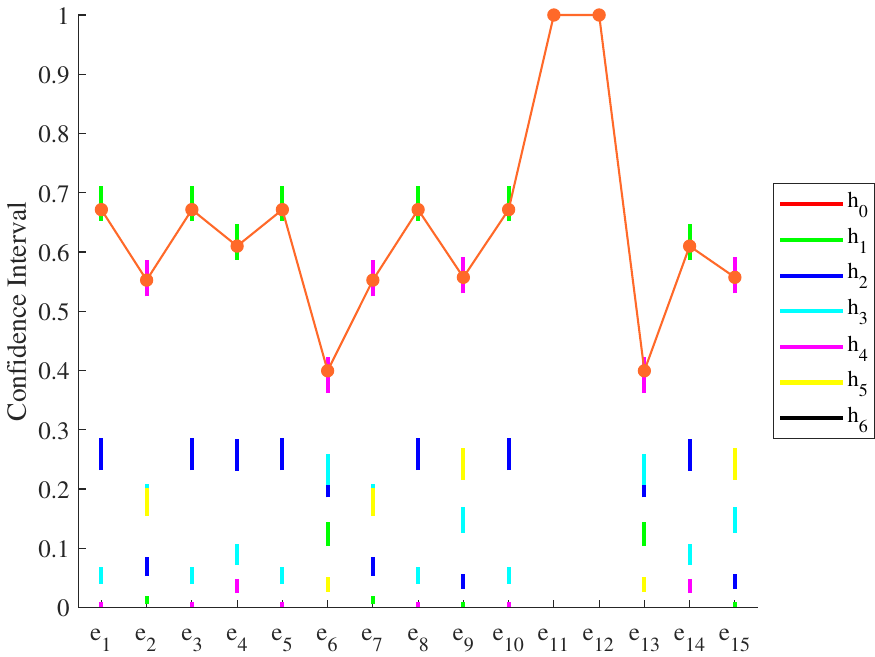}
\caption{Confidence Intervals and Maximum Representative Values of Each Agent after Monte Varlo Tests.}
\label{proposed_HK_same_proposed0.21}
\end{figure}
\begin{figure}[h]
\centering
\includegraphics[width=3.5in]{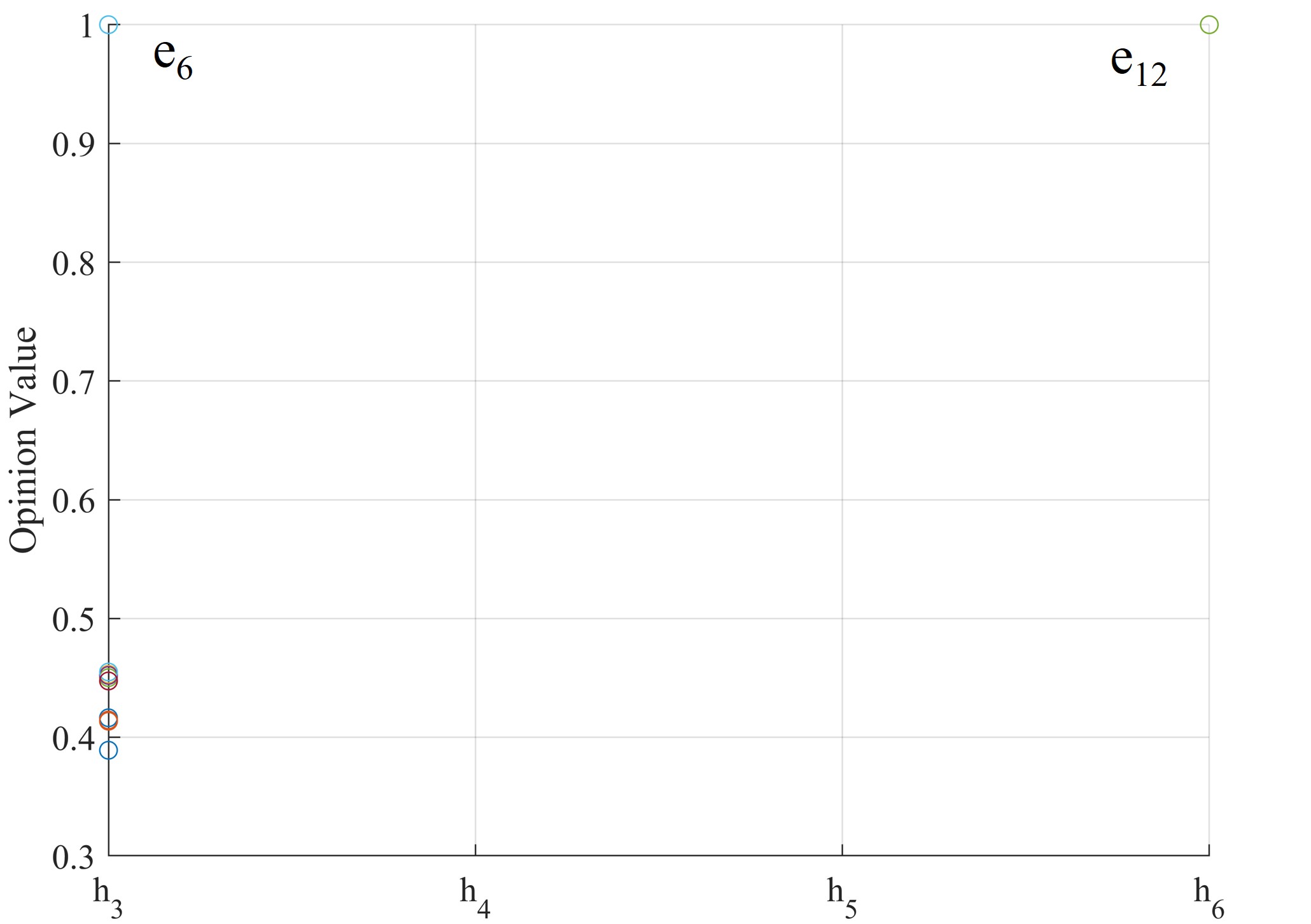}
\caption{Opinion Results Based on PRRLEM-HEHK.}
\label{HKdifferent_proposed}
\end{figure}

\begin{figure*}[h]
\centering
\includegraphics[width=6.6in]{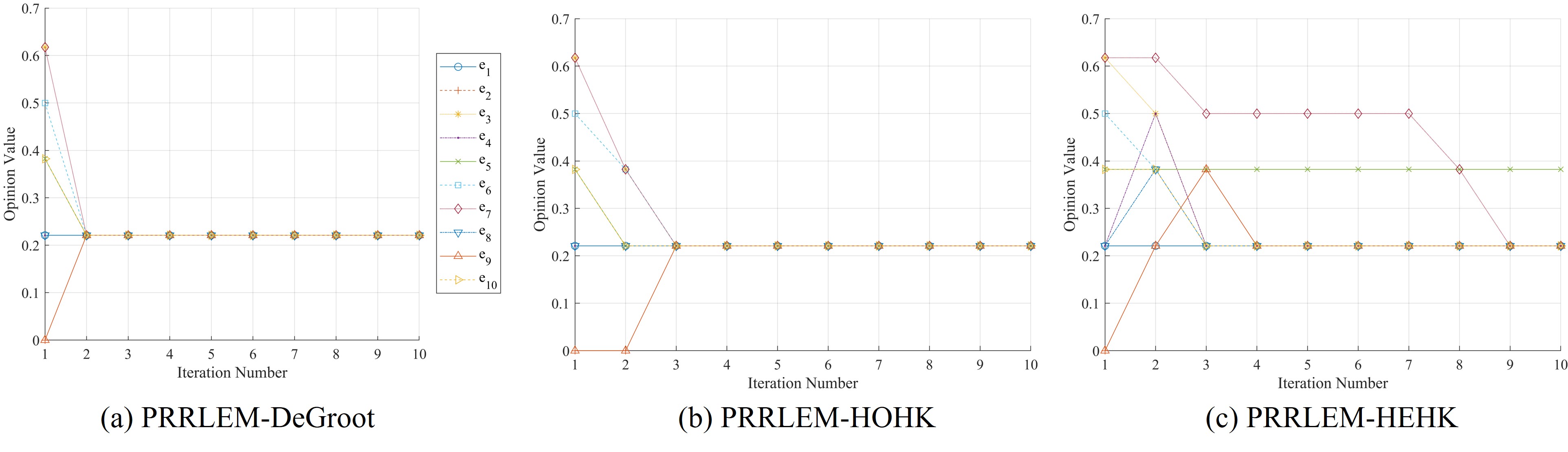}
\caption{Results Based on the Three Proposed Models for the Example about Space Situational Awareness.}
\label{space}
\end{figure*}

\begin{figure*}[h]
\centering
\includegraphics[width=7.1in]{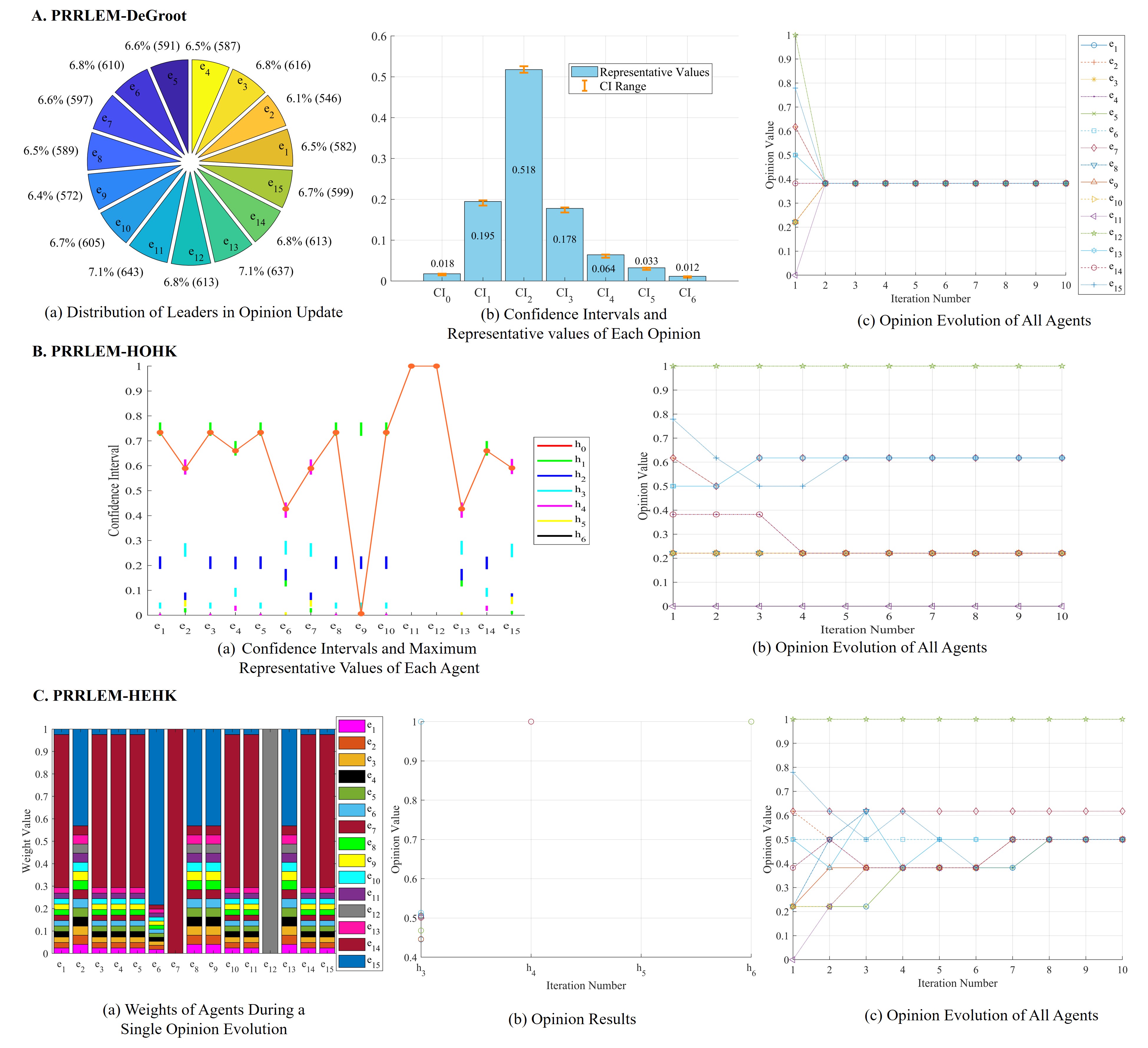}
\caption{Results Based on the Proposed Three Models for Robustness Analysis.}
\label{Robustness}
\end{figure*}

\begin{figure*}[h]
\centering
\includegraphics[width=5.5in]{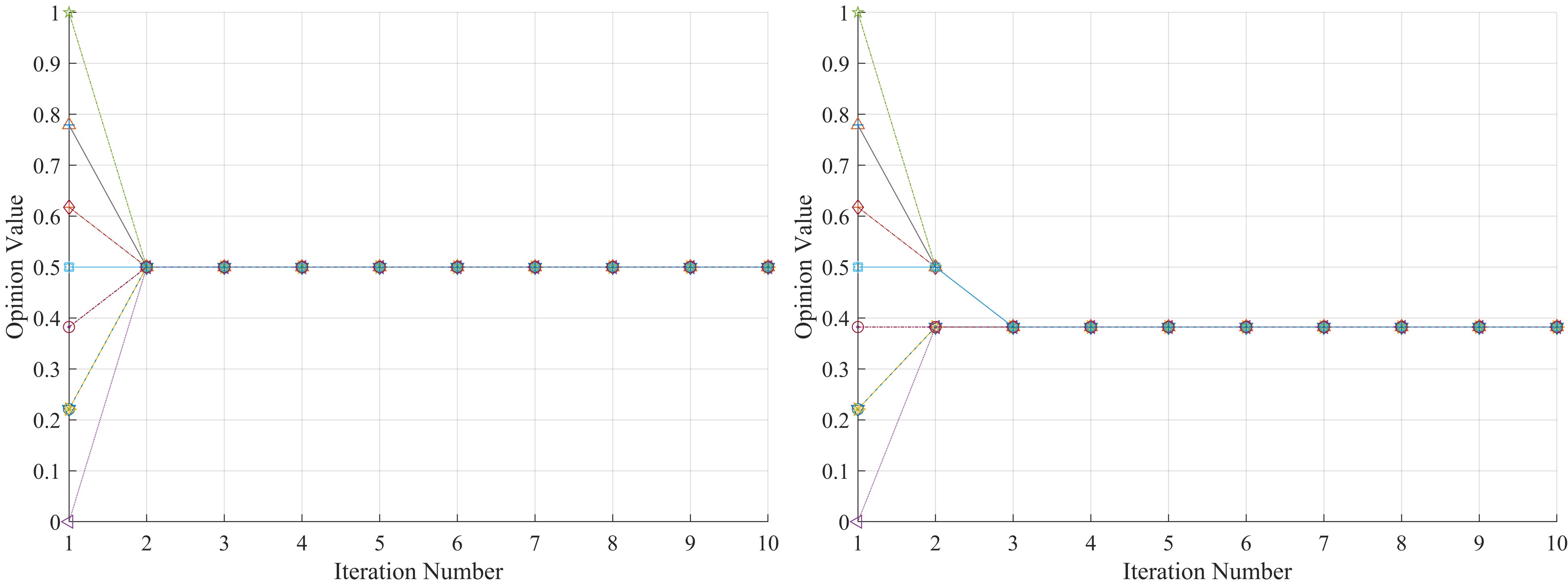}
\caption{Results Based on the DeGroot model.}
\label{DG_compare}
\end{figure*}
\begin{figure*}[h]
\centering
\includegraphics[width=5.8in]{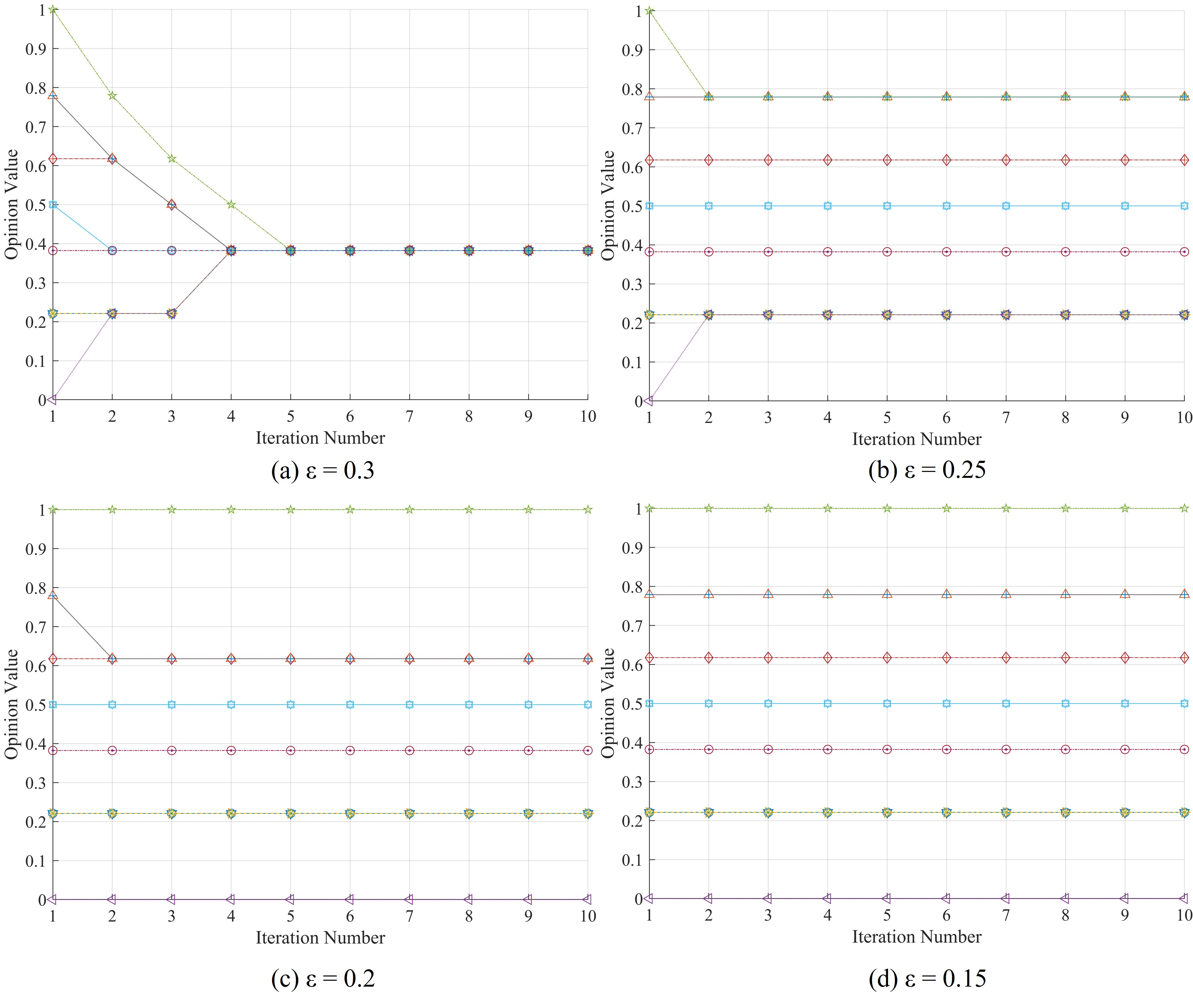}
\caption{Results Based on Homogeneous HK Bounded Confidence Model.}
\label{HKsame_compare}
\end{figure*}
\begin{figure*}[h]
\centering
\includegraphics[width=6.0in]{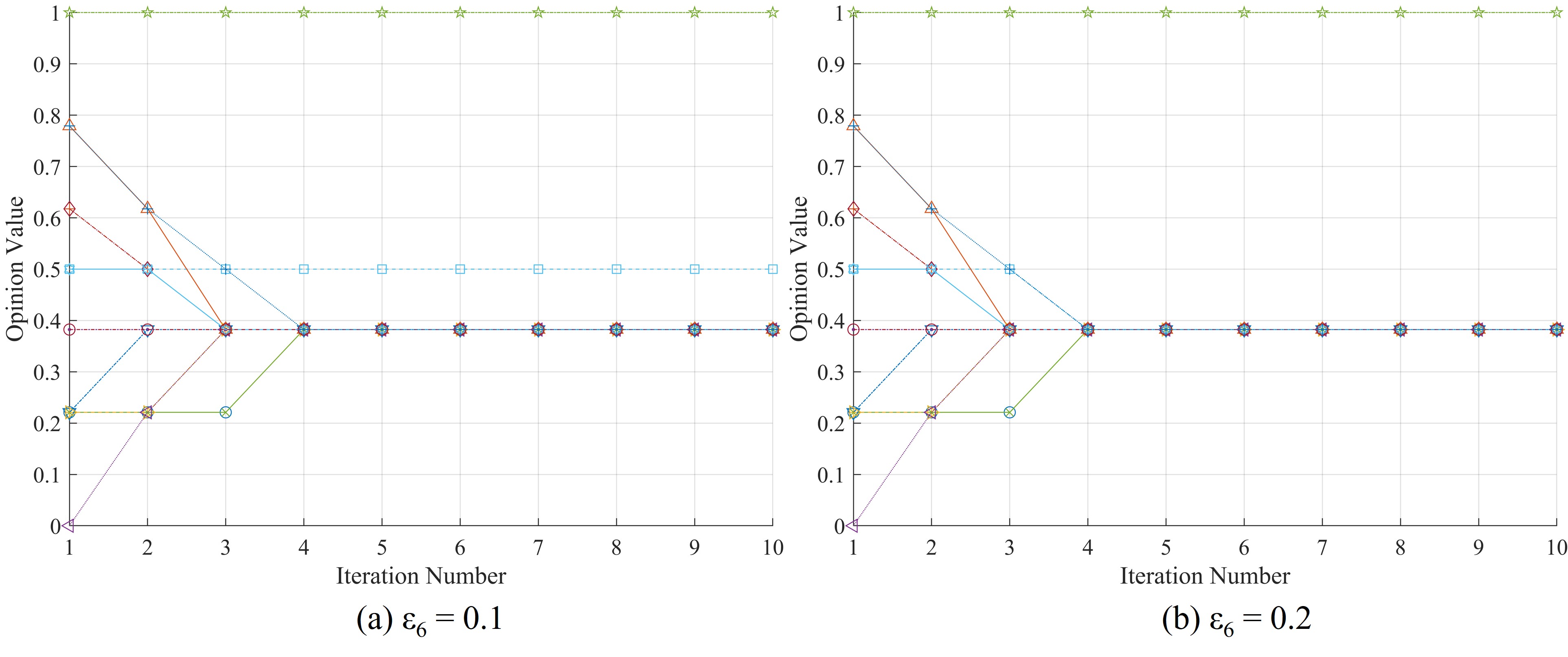}
\caption{Results Based on Traditional Heterogeneous HK Bounded Confidence Model.}
\label{HKdifferent_original}
\end{figure*}
\section{Illustrate Examples and Analysis}
\par In this section, we employ illustrative examples to demonstrate the effectiveness of the proposed models. We aim to illustrate the applicability of our theoretical framework in real-world scenarios through these examples.
\subsection{Example 1 for PRRLEM-DeGroot}
\par \emph{Step 1: Initialization}
\par The simulation involves \( N = 15 \) agents. Considering that the proposed models do not involve complex computations, the number of the Monte Carlo method is \( 1000 \) times (\( M = 1000 \)). For presentation of the results, we set the number of opinion update to nine (\( T = 9 \)) in each Monte Carlo test. Opinions are represented by a finite and completely ordered set of linguistic terms \( H = \{ h_0, h_1, \ldots, h_6 \} \) (\( \Phi = 3 \)). In \cite{2010Improved}, the authors state that \( a \) in ($\ref{eq:trans}$) is most likely to belong to [1.36, 1.4]. Therefore, we set \( a \) to \( 1.37 \). When \( t = 1 \), each agent offers an initial opinion, as follows: \( \kappa_{e_1}(1) = h_1 \), \( \kappa_{e_2}(1) = h_4 \), \( \kappa_{e_3}(1) = h_1 \), \( \kappa_{e_4}(1) = h_2 \), \( \kappa_{e_5}(1) = h_1 \), \( \kappa_{e_6}(1) = h_3 \), \( \kappa_{e_7}(1) = h_4 \), \( \kappa_{e_8}(1) = h_1 \), \( \kappa_{e_9}(1) = h_5 \), \( \kappa_{e_{10}}(1) = h_1 \), \( \kappa_{e_{11}}(1) = h_0 \), \( \kappa_{e_{12}}(1) = h_6 \), \( \kappa_{e_{13}}(1) = h_3 \), \( \kappa_{e_{14}}(1) = h_2 \), and \( \kappa_{e_{15}}(1) = h_5 \). 
\par \emph{Step 2: Opinion Conversion}
 \par Based on (\ref{eq:trans}), the corresponding numerical values of the linguistic terms can be obtained, as shown in Table \uppercase\expandafter{\romannumeral1}.
\begin{table}[!h]
\centering
\caption{Numerical Values of \\Linguistic Terms When $a = 1.37$.}
\setlength{\tabcolsep}{1.2mm}
\renewcommand\arraystretch{1.5}
\begin{tabular}{c|ccccccccccccccc}
\bottomrule
$h_\xi$&$\xi=0$&$\xi=1$&$\xi=2$&$\xi=3$&$\xi=4$&$\xi=5$
&$\xi=6$\\
\hline
$\theta_\xi$&0&0.221 &0.382 &0.500 &0.618 &0.780 &1 
 \\
\toprule
\end{tabular}
 \end{table} 
\par \emph{Step 3: Weight Allocation}
\par Table \uppercase\expandafter{\romannumeral2} shows the distribution of leaders in opinion updates. It can be seen that all 15 agents have taken on the role of leader in the $MT=9000$ events. The most frequent leader is $e_{13}$, accounting for 7.0\% (633), while the least frequent is $e_3$ with 6.3\% (568). The rest of the agents' frequencies are distributed between 6.4\% and 6.9\%, indicating a relatively balanced distribution pattern. This suggests that the proposed model promotes a balance among agents, which is essential for fair and effective decision-making.
\begin{table}[!h]
\centering
\caption{Distribution of Leaders in Opinion Update.}
\scriptsize
\setlength{\tabcolsep}{1.1mm}
\renewcommand\arraystretch{1.5}
\begin{tabular}{c|ccccccccccccccc}
\bottomrule
 &$e_1$&$e_2$&$e_3$&$e_4$&$e_5$\\
\hline
Percentage/Times & 6.5\%/(586) & 6.7\%/(604) &6.8\%/(610) & 6.3\%/(568) & 6.9\%/(623) 
 \\ \hline
  &$e_6$&$e_7$&$e_8$&$e_9$&$e_{10}$\\
\hline
 Percentage/Times& 6.8\%/(608) & 6.8\%/(608) &6.8\%/(611) & 6.4\%/(579) & 6.7\%/(607) 
 \\\hline
   &$e_{11}$&$e_{12}$&$e_{13}$&$e_{14}$&$e_{15}$\\
\hline
Percentage/Times & 6.6\%/(594) & 6.4\%/(576) &7.0\%/(633) & 6.5\%/(584) & 6.8\%/(609) 
 \\
\toprule
\end{tabular}
 \end{table}  
\par \emph{Step 4: Opinion Update}
\par Based on (\ref{eq:update}) and (\ref{eq:linguistic}), the agents update the opinions and derive the final results. Fig. \ref{updateopi} displays the opinion update process of all agents during a particular Monte Carlo test. It can be observed that the system reaches a consensus on $h_2$.

\par \emph{Step 5: Iterative Simulation and Result Recording}
\par After conducting 1000 Monte Carlo tests, the statistical results of each agent's opinions after each test are shown in Table \uppercase\expandafter{\romannumeral3}. The sum of the times of each agent's final opinions is 1000. The most frequent opinions observed are $h_2$ and $h_3$, with 5220 and 4830 times, while the least frequent linguistic terms are for $h_0$ and $h_6$, each with 285 times.
\begin{table}[!h]
\centering
\caption{Distribution of Agents' Opinions \\in Monte Carlo Tests.}
\setlength{\tabcolsep}{1.5mm}
\renewcommand\arraystretch{1.5}
\begin{tabular}{c|ccccccc|cccccccc}
\bottomrule
&$h_0$&$h_1$&$h_2$&$h_3$&$h_4$&$h_5$
&$h_6$&Total\\
\hline
$e_1$&19 &129 &348 &322 &106 &57 &19&1000 \\
$e_2$&19 &129 &348 &322 &106 &57 &19&1000 \\
$\vdots$& $\vdots$&$\vdots$ &$\vdots$  &$\vdots$ &$\vdots$  &$\vdots$  &$\vdots$&$\vdots$ \\
$e_{15}$&19 &129 &348 &322 &106 &57 &19&1000 \\
\hline
Total&285 &1935 &5220  &4830 & 1590 & 1425 & 285&15000 \\
\toprule
\end{tabular}
 \end{table} 

\par \emph{Step 6: Monte Carlo Analysis}
\par Utilize (\ref{eq:Monte}), the confidence intervals of linguistic terms are: $CI_0=[0.017,0.021]$, $CI_1=[0.124,0.134]$, $CI_2=[0.340,0.356]$, $CI_3=[0.315,0.330]$, $CI_4=[0.101,0.111]$, $CI_5=[0.053,0.061]$, and $CI_6=[0.017,0.021]$.

\par \emph{Step 7: Confidence Intervals Ranking and Results Determination}
\par Based on (\ref{eq:GRRV}), the corresponding representative values are: $Rep(CI_0)=0.021$, $Rep(CI_1)=0.133$, $Rep(CI_2)=0.350$, $Rep(CI_3)=0.325$, $Rep(CI_4)=0.110$, $Rep(CI_5)=0.060$, and $Rep(CI_6)=0.021$. The bar chart is shown as Fig. \ref{proposed_DG} to display the results visually. It can be seen that the values of $Rep(CI_2)$ and $Rep(CI_3)$ are very close, which also indicates that the model proposed in this paper can not only represent opinion evolution like the traditional DeGroot model but also quantify the differences of each opinion. Further comparisons with the traditional method will be conducted later.

\subsection{Example 2 for PRRLEM-HOHK}
\par Continuing with \emph{Example 1 for PRRLEM-DeGroot}, all initial settings remain unchanged. All agents have the same threshold $\varepsilon=0.21$. \emph{Step 1} and \emph{Step 2} remain unchanged. 
\par \emph{Step 3: Confidence Set Determination}
\par According to (\ref{eq:confidence_set}), the confidence set of each agent is determined, and it continues to change during the process of opinion evolution.
\par \emph{Step 4: Weight Allocation}
\par (\ref{eq:Sheep_opinion}) and (\ref{eq:theta}) are adopted to determine the weight vector of agents. 

\par \emph{Step 5: Opinion Update}
\par Based on (\ref{eq:update}) and (\ref{eq:linguistic}), the agents update the opinions.
\par \emph{Step 6: Iterative Simulation and Result Recording}
\par After Monte Carlo tests, the distribution of agents' opinions is shown in Table \uppercase\expandafter{\romannumeral4}. Two special cases are that for $e_{11}$ and $e_{12}$, the number of occurrences of $h_0$ and $h_6$ is both 1000. This is actually due to the threshold setting; the confidence sets of these two agents only contain themselves.
\begin{table}[!h]
\centering
\caption{Distribution of Agents' Opinions \\in Monte Carlo Tests.}
\setlength{\tabcolsep}{1.5mm}
\renewcommand\arraystretch{1.2}
\begin{tabular}{c|ccccccc|cccccccc}
\bottomrule
&$h_0$&$h_1$&$h_2$&$h_3$&$h_4$&$h_5$
&$h_6$&Total\\
\hline
$e_1$& 0&682 &259  &54  &5  &0 &0  &1000 \\
$e_2$&0 &12 &69  &185  &556  &178 & 0&1000 \\
$e_3$&0&682 &259  &54  &5  &0 &0& 1000  \\
$e_4$&0 & 617& 258 &89  &36  &0 &0& 1000 \\
$e_5$& 0&682 &259  &54  &5  &0 &0& 1000  \\
$e_6$&0 &124 &211  &233  &393  &39  &0& 1000\\
$e_7$& 0 &12 &69  &185  &556  &178 & 0& 1000\\
$e_8$& 0&682 &259  &54  &5  &0 &0&   1000\\
$e_9$&0 &5 &44  &148  &561  &242 &0&  1000\\
$e_{10}$& 0&682 &259  &54  &5  &0 &0&   1000\\
$e_{11}$&1000 &0 &0  &0  &0  &0  &0&1000 \\
$e_{12}$& 0&0 &0  & 0 &0  &0 & 1000&1000 \\
$e_{13}$& 0 &124 &211  &233  &393  &39  &0 &  1000\\
$e_{14}$& 0 & 617& 258 &89  &36  &0 &0& 1000 \\
$e_{15}$& 0 &5 &44  &148  &561  &242 &0&          1000\\
\hline
Total&1000  &4926  &2459  &  1580 & 3117  & 918  & 1000 & 15000 \\
\toprule
\end{tabular}
\end{table} 
\par \emph{Step 7: Monte Carlo Analysis}
\par The confidence intervals of each opinion and the corresponidng representative values are shown in Fig. \ref{proposed_HK_same_proposed0.21}.
\par \emph{Step 8: Confidence Intervals Ranking and Results Determination}
\par In Fig. \ref{proposed_HK_same_proposed0.21}, it can be seen that the agents $e_1$, $e_3$, $e_4$, $e_5$, $e_8$, $e_{10}$, and $e_{14}$ reach the same opinion $h_1$. Meanwhile, $e_2$, $e_6$, $e_7$, $e_9$, $e_{13}$, and $e_{15}$ reach $h_4$. $e_{11}$ and $e_{12}$ still maintain the original opinions, that is, $h_0$ and $h_6$.

\subsection{Example 3 for PRRLEM-HEHK}
\par Continuing with \emph{Example 2 for PRRLEM-HOHK}, all initial settings remain unchanged. The threshold values of agents are: $\varepsilon_1 = 0.2$, $\varepsilon_2 = 0.5$, $\varepsilon_3 = 0.3$, $\varepsilon_4 = 0.4$, $\varepsilon_5 = 0.2$, $\varepsilon_6 = 0.1$, $\varepsilon_7 = 0.9$, $\varepsilon_8 = 0.6$, $\varepsilon_9 = 0.5$, $\varepsilon_{10} = 0.3$, $\varepsilon_{11} = 0.3$, $\varepsilon_{12} = 0.1$, $\varepsilon_{13} = 0.8$, $\varepsilon_{14} = 0.4$, and $\varepsilon_{15} = 0.2$. After Monte Carlo tests, the final results are displayed in Fig. \ref{HKdifferent_proposed}. As can be seen, except for $e_{12}$, all other agents have reached a consensus on opinion $h_3$. This is because the threshold value $\varepsilon_{12}$ is minimal, only 0.1, resulting in no other agents within this agent's confidence set. The situation for $e_6$ is similar.

\subsection{Real-World Example about Space Situational Awareness}
\par In the current space environment, the continuous increase in human space activities has led to a significant growth in space debris and the number of spacecraft in orbit. The growing amount of space debris has crowded the orbital resources, posing a threat to the safe operation of spacecraft. Impacts from space debris of centimeter size or more significant can lead to punctures in spacecraft or even disintegration until wholly damaged. Impacts from space debris smaller than a centimeter can cause partial malfunction or failure of spacecraft, and damage to critical components may also lead to the failure of the entire satellite. This trend has threatened the regular operation of satellites and space stations. Facing this challenge, this study utilizes the opinion dynamics to invite space enthusiasts and researchers to evaluate the future development trends of space situational awareness \cite{wang2022research}. 
\par We invite ten agents, and the linguistic term set is \( H = \{ h_0, h_1, \ldots, h_6 \} \). \( M = 1000 \), \( T = 9 \), and \( a=1.37 \). The initial opinions of agents are: \( \kappa_{e_1}(1) = h_1 \), \( \kappa_{e_2}(1) = h_0 \), \( \kappa_{e_3}(1) = h_4 \), \( \kappa_{e_4}(1) = h_1 \), \( \kappa_{e_5}(1) = h_2 \), \( \kappa_{e_6}(1) = h_3 \), \( \kappa_{e_7}(1) = h_4 \), \( \kappa_{e_8}(1) = h_1 \), \( \kappa_{e_9}(1) = h_0 \), \( \kappa_{e_{10}}(1) = h_2 \). For PRRLEM-HOHK, the threshold value is 0.4. For PRRLEM-HEHK, the threshold values of agents are: $\varepsilon_1 = 0.1$, $\varepsilon_2 = 0.3$, $\varepsilon_3 = 0.7$, $\varepsilon_4 = 0.5$, $\varepsilon_5 = 0.1$, $\varepsilon_6 = 0.3$, $\varepsilon_7 = 0.2$, $\varepsilon_8 = 0.3$, $\varepsilon_9 = 0.3$, and $\varepsilon_{10} = 0.3$. After calculation, the final results are shown in Fig. \ref{space}. 

\par The results from the three models indicate that the future development trend of the space environment may be severe, necessitating research into space debris cleanup and collision avoidance. This result is consistent with the current situation. Many countries are promoting space debris cleanup plans and conducting in-depth research on space situational awareness. The European Space Agency (ESA) plans to deploy an innovative technology in 2025 with the ClearSpace-1 mission, which aims to capture and remove space debris using a clamping mechanism. Meanwhile, the Japan Aerospace Exploration Agency (JAXA) is utilizing electrodynamic tether technology in its KITE project, targeting the removal of low-orbit debris by drawing it into the atmosphere for destruction, thereby mitigating the space debris issue. Additionally, the United States Space Surveillance Network, with its global network of radars and telescopes, not only enhances the capability of space situational awareness but also helps prevent potential collisions.

\section{Robustness Analysis and Comparison Analysis}
\par To verify the superiority of the proposed models, a detailed robustness analysis and comparative analysis with other methods are conducted. 
\subsection{Robustness Analysis}
\par Conducting a robustness analysis of models is crucial, especially in complex environments where models may face various abnormal issues and extreme conditions. By intentionally introducing biased processing and deviations from normal data values to simulate potential data abnormalities or interference, we can effectively test the performance of models under non-ideal conditions.
\par We test the robustness of PRRLEM-DeGroot and PRRLEM-HOHK models by replacing the initial opinion \( \kappa_{e_9}(1) = h_5 \) in \emph{Example 1 for PRRLEM-DeGroot} and \emph{Example 2 for PRRLEM-HOHK}  with \( \kappa_{e_9}(1) = h_1 \). The rest remains the same. For \emph{Example 3 for PRRLEM-HEHK}, we not only replace \( \kappa_{e_9}(1) = h_5 \) with \( \kappa_{e_9}(1) = h_1 \) but also replace $\varepsilon_7 = 0.9$ with $\varepsilon_7 = 0.1$ to test the effect of the threshold value on the results.
\par As can be seen in Fig. \ref{Robustness}, although \( \kappa_{e_9}(1) \) shifts significantly from the very positive $h_5$ to $h_1$, the final results do not change, which demonstrates the robustness of both models. For PRRLEM-HEHK, even if we change both the threshold $\varepsilon_7$ and \( \kappa_{e_9}(1) \), the only difference is the presence of $h_4$, which is mainly due to the adjustment of $\varepsilon_7$ from 0.9 to the extreme 0.1. The other results remain the same as before. Therefore, the three models proposed in this paper show strong anti-interference and robustness.
\subsection{Comparison Analysis}
\subsubsection{Comparison with Traditional DeGroot Model}
\par The initial setup is consistent with \emph{Example 1 for PRRLEM-DeGroot}. Fig. \ref{DG_compare} shows the results obtained by the traditional DeGroot model. For the left picture, the weights of agents in the opinion evolution is $\Omega=\lbrace\omega_1,\omega_2,\cdots,\omega_{15}\rbrace=\lbrace\frac{1}{15}, \frac{1}{15},\cdots,\frac{1}{15}\rbrace$. For the right picture, the agents' weight vector when $e_i$ updates the opinion is based on the distance between opinions $\omega_j = \frac{e^{-\vert D_{ij}\vert}}{\sum_{k=1}^{15} e^{-\vert D_{ik}\vert}}$. Obviously, the final results are quite different in both cases. In the first case all agents reach $h_3$, while in the second case all agents reach $h_2$. It indicates that traditional model is extremely sensitive to the choice of parameters; slight differences can result in significant variances in the final results.

\par Unlike the traditional DeGroot model, the proposed model introduces a randomness principle that enables dynamic leadership transfer among agents. This unique random leadership model significantly enhances the system's adaptability to external changes in decision-making across variable environments. Our proposed model can also present these close opinions and assign an exact value to each opinion, enabling the experts to form decisions more comprehensively. Besides, the results highlight an essential advantage: robustness. Our model shows stability that does not depend on a specific weight vector, ensuring that uncertainties in the decision-making process are effectively managed.
\begin{table*}[!h]
\centering
\scriptsize
\caption{Representative Values of Agents on Opinions Based on PRRLEM-HOHK When $\varepsilon$ Takes Different Values}
\setlength{\tabcolsep}{2.2mm}
\renewcommand\arraystretch{1.0}
{
\begin{tabular}{cccccccc|ccccccccccc}
\bottomrule
$\varepsilon=0.3$&  &  &  &  &   & &  &$\varepsilon=0.25$  &\\\hline
&$h_0$&$h_1$& \textcolor{blue}{$h_2$}& $h_3$& $h_4$ & $h_5$& $h_6$&&$h_0$&\textcolor{blue}{$h_1$}& $h_2$& \textcolor{blue}{$h_3$}& $h_4$ & \textcolor{blue}{$h_5$}& $h_6$\\
\hline
$e_1$&0.034 &0.280  & \textcolor{blue}{0.420}  &0.246  &  0.072 &0.005  &0  & $e_1$&0.055 & \textcolor{blue}{0.633} & 0.245 & 0.082 & 0.025 & 0  & 0 \\
$e_2$&0.003  &0.145  &\textcolor{blue}{0.406}   &0.257  &0.167   &0.087  &0.005  & $e_2$& 0 & 0.111 & 0.221 & \textcolor{blue}{0.339} & 0.230  & 0.167 & 0.006\\
$e_3$&0.034  &0.280  &\textcolor{blue}{0.420}  &0.246  &   0.072&0.005  &0  & $e_3$& 0.055 & \textcolor{blue}{0.633} & 0.245 & 0.082 &   0.025& 0 & 0\\
$e_4$&0.011  &0.256   &\textcolor{blue}{0.422}  &0.254  &0.100   &0.015  &0  &$e_4$ & 0.016 & \textcolor{blue}{0.405} & 0.252 & 0.244 &   0.127& 0.016 & 0.003\\
$e_5$&0.034  &0.280  &\textcolor{blue}{0.420}  &0.246  &   0.072& 0.005 &  0& $e_5$& 0.055 &\textcolor{blue}{0.633} & 0.245 & 0.082 &   0.025& 0 & 0\\
$e_6$& 0.009 & 0.226 & \textcolor{blue}{0.416} & 0.256 &   0.125& 0.030 & 0.003 & $e_6$& 0 & 0.143 & 0.248 & \textcolor{blue}{0.387} &    0.224& 0.058 & 0.006\\
$e_7$& 0.003 & 0.145 & \textcolor{blue}{0.406} & 0.257 &   0.167& 0.087 & 0.005 & $e_7$& 0 & 0.111 & 0.221 & \textcolor{blue}{0.339} &   0.230 & 0.167 &0.006\\
$e_8$& 0.034 & 0.280 & \textcolor{blue}{0.420} & 0.246 &   0.072& 0.005 & 0 & $e_8$& 0.055 & \textcolor{blue}{0.633} & 0.245 & 0.082 &  0.025 & 0 &0\\
$e_9$& 0 & 0.104 &  \textcolor{blue}{0.352}& 0.252 &  0.182  &0.150   & 0.039 & $e_9$& 0 & 0.022 & 0.071 & 0.142 &   0.226& \textcolor{blue}{0.532} & 0.068\\
$e_{10}$& 0.034 & 0.280 & \textcolor{blue}{0.420} & 0.246 & 0.072  & 0.005 & 0 & $e_{10}$& 0.055 &   \textcolor{blue}{0.633}& 0.245 & 0.082 &  0.025 & 0 & 0\\
$e_{11}$& 0.049 & 0.280  & \textcolor{blue}{0.416} & 0.241 & 0.069  & 0.005 & 0 & $e_{11}$& 0.067 & \textcolor{blue}{0.632} & 0.236 & 0.082 & 0.024  & 0 & 0\\
$e_{12}$& 0 & 0.078 & \textcolor{blue}{0.309} & 0.233 &   0.172& 0.155 & 0.139 & $e_{12}$& 0 & 0.017 & 0.063 & 0.126 &   0.223& \textcolor{blue}{0.542} &0.088\\
$e_{13}$& 0.009  & 0.226 & \textcolor{blue}{0.416} & 0.256 & 0.125  & 0.030 & 0.003 & $e_{13}$& 0 & 0.143 &  0.248 &  \textcolor{blue}{0.387}&  0.224 & 0.058  & 0.006\\
$e_{14}$&  0.011& 0.256 & \textcolor{blue}{0.422} & 0.254 & 0.100   & 0.015 & 0 & $e_{14}$& 0.016 & \textcolor{blue}{0.405} & 0.252 & 0.244 &  0.127 & 0.016 & 0.003\\
$e_{15}$& 0 & 0.104 &\textcolor{blue}{0.352}  & 0.252 &   0.182&  0.150 & 0.039 & $e_{15}$& 0 & 0.022 &  0.071 & 0.142 &   0.226 & \textcolor{blue}{0.532} &0.068\\
\hline
$\varepsilon=0.2$&  &  &  &  &   & &  &$\varepsilon=0.15$  &\\\hline
&\textcolor{blue}{$h_0$}&\textcolor{blue}{$h_1$}& $h_2$& $h_3$& \textcolor{blue}{$h_4$} & $h_5$& \textcolor{blue}{$h_6$}&&\textcolor{blue}{$h_0$}&\textcolor{blue}{$h_1$}& $h_2$& \textcolor{blue}{$h_3$}& $h_4$ & \textcolor{blue}{$h_5$}& \textcolor{blue}{$h_6$}\\
\hline
$e_1$& 0& \textcolor{blue}{0.660} & 0.281 & 0.068&0.012  & 0 & 0 & $e_1$& 0&\textcolor{blue}{1}  & 0& 0& 0 &0  &0  \\
$e_2$& 0& 0.016 & 0.082 &0.200 &  \textcolor{blue}{0.558} &0.190  &0  & $e_2$&0 & 0 & 0.124& \textcolor{blue}{0.683}& 0.212 &0  &0  \\
$e_3$& 0&\textcolor{blue}{0.660}  & 0.281 & 0.068& 0.012 &0 &0  & $e_3$& 0&  \textcolor{blue}{1}& 0& 0& 0 &0 & 0 \\
$e_4$& 0& \textcolor{blue}{0.590} & 0.278  & 0.111&0.052  & 0.005& 0 & $e_4$&0 & 0 &0.214 & \textcolor{blue}{0.683}& 0.122  & 0 & 0 \\
$e_5$& 0&\textcolor{blue}{0.660}  & 0.281 & 0.068& 0.012 & 0&0  & $e_5$& 0&  \textcolor{blue}{1}& 0&0 & 0 &0  & 0 \\
$e_6$& 0& 0.126 & 0.221 & 0.247& \textcolor{blue}{0.422} & 0.043& 0 & $e_6$&0 & 0 & 0.164& \textcolor{blue}{0.683}& 0.174 & 0 &  0\\
$e_7$&0 & 0.016 & 0.082  &0.200 &\textcolor{blue}{0.558}  &0.190 & 0 & $e_7$&0 & 0 &0.124 &\textcolor{blue}{0.683} & 0.212 & 0 &  0\\
$e_8$& 0&\textcolor{blue}{0.660}   & 0.281 & 0.068& 0.012 & 0& 0 & $e_8$& 0&  \textcolor{blue}{1}& 0&0 & 0 &  0&  0\\
$e_9$&0 & 0.005 &0.041  &0.161 & \textcolor{blue}{0.565} & 0.266& 0 & $e_9$&0 & 0 &0 & 0& 0 & \textcolor{blue}{1} & 0 \\
$e_{10}$& 0& \textcolor{blue}{0.660} &0.281  &  0.068& 0.012  & 0& 0 & $e_{10}$&0 & \textcolor{blue}{1} &0 & 0& 0 & 0 & 0 \\
$e_{11}$& \textcolor{blue}{1}& 0 &0  & 0& 0 & 0&  0& $e_{11}$& \textcolor{blue}{1}& 0 &0 & 0&  0& 0 &0  \\
$e_{12}$& 0& 0 & 0 & 0& 0 &0 &\textcolor{blue}{1}  & $e_{12}$&0 &0  &0& 0& 0 & 0 &\textcolor{blue}{1}  \\
$e_{13}$& 0& 0.126 & 0.221 &0.247 & \textcolor{blue}{0.422} & 0.043& 0 & $e_{13}$&0 & 0 &0.164 &\textcolor{blue}{0.683} & 0.174 & 0 & 0 \\
$e_{14}$& 0&\textcolor{blue}{0.590}  & 0.278 &0.111 &  0.052 &0.005 & 0 & $e_{14}$& 0& 0 & 0.214& \textcolor{blue}{0.683}& 0.122 &0  &  0\\
$e_{15}$& 0& 0.005 & 0.041 &  0.161& \textcolor{blue}{0.565} &0.266 & 0 & $e_{15}$&0 &0  & 0&0 & 0 & \textcolor{blue}{1} & 0 \\
\toprule
\end{tabular}
}
 \end{table*}

\subsubsection{Comparison with Traditional Homogeneous HK Bounded Confidence Model}
\par The initial setup is consistent with \emph{Example 2 for PRRLEM-HOHK}. The results based on the traditional homogeneous HK bounded confidence model are shown in Fig. \ref{HKsame_compare}. As the confidence value $\varepsilon$ decreases, the communication between agents reduces. When $\varepsilon$ reaches 0.15, the traditional model results in agents completely ceasing communication with those holding different opinions. The results based on the proposed PRRLEM-HOHK model is shown in Table \uppercase\expandafter{\romannumeral5}. The blue marks in the table represent the maximum value of each row. For example, the first blue mark 0.420 indicates the representative value of $e_1$'s confidence interval for these seven linguistic terms $h_0$, $h_1$, $h_2$, $h_3$, $h_4$, $h_5$, and $h_6$ after Monte Carlo test when $\varepsilon=0.3$. These agents reach the consensus $h_2$ when $\varepsilon=0.3$. We can see that when $\varepsilon=0.15$, the agents in our model still be exchanging opinions and form the final results $h_0$, $h_1$, $h_3$, $h_5$, and $h_6$.

\par In opinion dynamics research, a common challenge arises when agents have low trust in information that differs from their opinions, leading to a phenomenon known as \emph{echo chamber}. This results in group polarization, as agents are inclined to communicate only with those who share the same opinions, thus distrusting agents with different opinions. However, our model maintains the exchange of opinions even when agents' trust is very low. This innovation indicates our model's potential to reduce the effect of group polarization and slow the speed of forming echo chamber and shows its effectiveness in preserving information sharing.

\subsubsection{Comparison with Traditional Heterogeneous HK Bounded Confidence Model}
\par The initial setup is consistent with \emph{Example 3 for PRRLEM-HEHK}. The results based on the traditional heterogeneous HK bounded confidence model are shown in Fig. \ref{HKdifferent_original}(a). When $\varepsilon_6$ becomes 0.2 and the threshold values of other agents remain the same, the final opinions of agents are shown in Fig. \ref{HKdifferent_original}(b). As can be seen, simply due to a difference in the setting of one parameter, there are significant differences in the final results. However, based on the proposed algorithm, the final results remain consistent with Fig. \ref{HKdifferent_proposed}. Therefore, this comparison once again demonstrates that this paper's research can improve the robustness of the opinion dynamics model, which is crucial to reduce the influence of the initial conditions set in the opinion dynamics model and the interference existing in the information sharing on the final results.

\section{Conclusion}
\par In this paper, we have introduced three linguistic fuzzy information evolution algorithms for understanding how group information change. These models are based on the PRRLEM, showing us a new way to look at leadership and decision-making among agents. We use the Monte Carlo method and confidence intervals to derive better the variety of possible opinions rather than just one fixed result. Our simulations show that these new models can help groups share information better, make more diverse choices, and decide more effectively. By changing who leads at different times, we avoid depending too much on just one leader. This can make the group's decisions stronger and less likely to fail because of one agent's mistake. Also, by using confidence intervals to express opinions and GRRV to rank these intervals, we can deal with the uncertainty and randomness that often challenge the traditional information evolution models.
\par To sum up, our models offer a novel perspective on how information forms and changes within groups, drawing lessons from nature and using mathematical methods to handle uncertainty. These models could be helpful in analyzing social media, planning marketing strategies, and building agreement among people. In the future, we will study the effect of noise on the proposed models.

\bibliography{mybibfile} 

\begin{thebibliography}{10}
\providecommand{\url}[1]{#1}
\csname url@samestyle\endcsname
\providecommand{\newblock}{\relax}
\providecommand{\bibinfo}[2]{#2}
\providecommand{\BIBentrySTDinterwordspacing}{\spaceskip=0pt\relax}
\providecommand{\BIBentryALTinterwordstretchfactor}{4}
\providecommand{\BIBentryALTinterwordspacing}{\spaceskip=\fontdimen2\font plus
\BIBentryALTinterwordstretchfactor\fontdimen3\font minus
  \fontdimen4\font\relax}
\providecommand{\BIBforeignlanguage}[2]{{%
\expandafter\ifx\csname l@#1\endcsname\relax
\typeout{** WARNING: IEEEtran.bst: No hyphenation pattern has been}%
\typeout{** loaded for the language `#1'. Using the pattern for}%
\typeout{** the default language instead.}%
\else
\language=\csname l@#1\endcsname
\fi
#2}}
\providecommand{\BIBdecl}{\relax}
\BIBdecl

\bibitem{10654535}
P.~Liu, X.~Wang, Y.~Fu, and P.~Wang, ``Graph model for conflict resolution for
  mixed-stability combinatorial foresight based on the combination of regret
  theory and vikor method,'' \emph{IEEE Transactions on Fuzzy Systems}, pp.
  1--14, 2024.

\bibitem{9806378}
P.~Liu, Y.~Li, and P.~Wang, ``Opinion dynamics and minimum adjustment-driven
  consensus model for multi-criteria large-scale group decision making under a
  novel social trust propagation mechanism,'' \emph{IEEE Transactions on Fuzzy
  Systems}, vol.~31, no.~1, pp. 307--321, 2023.

\bibitem{9780145}
S.~Miao and H.~Su, ``Consensus of matrix-weighted hybrid multiagent systems,''
  \emph{IEEE Transactions on Cybernetics}, vol.~53, no.~1, pp. 668--678, 2023.

\bibitem{muslim2024mass}
R.~Muslim, R.~A. Nqz, and M.~A. Khalif, ``Mass media and its impact on opinion
  dynamics of the nonlinear q-voter model,'' \emph{Physica A: Statistical
  Mechanics and its Applications}, vol. 633, p. 129358, 2024.

\bibitem{liang2016fusion}
H.~Liang, C.-C. Li, Y.~Dong, and Y.~Jiang, ``The fusion process of interval
  opinions based on the dynamic bounded confidence,'' \emph{Information
  Fusion}, vol.~29, pp. 112--119, 2016.

\bibitem{jiao2021active}
Y.~Jiao and Y.~Li, ``An active opinion dynamics model: The gap between the
  voting result and group opinion,'' \emph{Information Fusion}, vol.~65, pp.
  128--146, 2021.

\bibitem{benatti2020opinion}
A.~Benatti, H.~F. de~Arruda, F.~N. Silva, C.~H. Comin, and
  L.~da~Fontoura~Costa, ``Opinion diversity and social bubbles in adaptive
  sznajd networks,'' \emph{Journal of Statistical Mechanics: Theory and
  Experiment}, vol. 2020, no.~2, p. 023407, 2020.

\bibitem{forgerini2023directed}
F.~L. Forgerini, N.~Crokidakis, and M.~A. Carvalho, ``Directed propaganda in
  the majority-rule model,'' \emph{arXiv preprint arXiv:2309.13026}, 2023.

\bibitem{9028187}
Y.~Zou and Z.~Meng, ``Targeted bipartite consensus of opinion dynamics in
  social networks with credibility intervals,'' \emph{IEEE Transactions on
  Cybernetics}, vol.~52, no.~1, pp. 372--383, 2022.

\bibitem{jiang2018predicting}
L.~Jiang, J.~Liu, D.~Zhou, Q.~Zhou, X.~Yang, and G.~Yu, ``Predicting the
  evolution of hot topics: A solution based on the online opinion dynamics
  model in social network,'' \emph{IEEE Transactions on Systems, Man, and
  Cybernetics: Systems}, vol.~50, no.~10, pp. 3828--3840, 2018.

\bibitem{su2021noise}
W.~Su, X.~Chen, Y.~Yu, and G.~Chen, ``Noise-based control of opinion
  dynamics,'' \emph{IEEE Transactions on Automatic Control}, vol.~67, no.~6,
  pp. 3134--3140, 2021.

\bibitem{9233959}
S.~Ke and W.~Liu, ``Consistency of multiagent distributed generative
  adversarial networks,'' \emph{IEEE Transactions on Cybernetics}, vol.~52,
  no.~6, pp. 4886--4896, 2022.

\bibitem{10526446}
S.~Liu, S.-K. Oh, W.~Pedrycz, B.~Yang, L.~Wang, and Z.~Peng, ``Scinn: Semantic
  concept-based inference neural networks with explainable and deep fuzzy
  structure,'' \emph{IEEE Transactions on Fuzzy Systems}, vol.~32, no.~7, pp.
  4133--4147, 2024.

\bibitem{10495183}
Y.~Teng, K.~Wu, and J.~Liu, ``Causal discovery from abundant but noisy fuzzy
  cognitive map set,'' \emph{IEEE Transactions on Fuzzy Systems}, vol.~32,
  no.~7, pp. 3992--4003, 2024.

\bibitem{LIU201534}
S.~Liu, C.~Jiang, Z.~Lin, Y.~Ding, R.~Duan, and Z.~Xu, ``Identifying effective
  influencers based on trust for electronic word-of-mouth marketing: A
  domain-aware approach,'' \emph{Information Sciences}, vol. 306, pp. 34--52,
  2015.

\bibitem{almeida2019optimal}
R.~Almeida, A.~B. Malinowska, and T.~Odzijewicz, ``Optimal leader--follower
  control for the fractional opinion formation model,'' \emph{Journal of
  Optimization Theory and Applications}, vol. 182, pp. 1171--1185, 2019.

\bibitem{dabarera2016consensus}
R.~Dabarera, K.~Premaratne, M.~N. Murthi, and D.~Sarkar, ``Consensus in the
  presence of multiple opinion leaders: Effect of bounded confidence,''
  \emph{IEEE Transactions on Signal and Information Processing over Networks},
  vol.~2, no.~3, pp. 336--349, 2016.

\bibitem{9711916}
Z.~Zhao, L.~Shi, T.~Li, J.~Shao, and Y.~Cheng, ``Opinion dynamics of social
  networks with intermittent-influence leaders,'' \emph{IEEE Transactions on
  Computational Social Systems}, vol.~10, no.~3, pp. 1073--1082, 2023.

\bibitem{gomez2022intermittent}
L.~G{\'o}mez-Nava, R.~Bon, and F.~Peruani, ``Intermittent collective motion in
  sheep results from alternating the role of leader and follower,''
  \emph{Nature Physics}, vol.~18, no.~12, pp. 1494--1501, 2022.

\bibitem{9204382}
X.~Wu, H.~Liao, and W.~Pedrycz, ``Probabilistic linguistic term set with
  interval uncertainty,'' \emph{IEEE Transactions on Fuzzy Systems}, vol.~29,
  no.~11, pp. 3532--3545, 2021.

\bibitem{JIA2023105610}
Q.~Jia, J.~Hu, W.~Zhang, S.~Zhai, and Z.~Li, ``A new situation assessment
  method for aerial targets based on linguistic fuzzy sets and trapezium
  clouds,'' \emph{Engineering Applications of Artificial Intelligence}, vol.
  117, p. 105610, 2023.

\bibitem{zhou2020two}
Q.~Zhou, Z.~Wu, A.~H. Altalhi, and F.~Herrera, ``A two-step communication
  opinion dynamics model with self-persistence and influence index for social
  networks based on the degroot model,'' \emph{Information Sciences}, vol. 519,
  pp. 363--381, 2020.

\bibitem{wu2022mixed}
Z.~Wu, Q.~Zhou, Y.~Dong, J.~Xu, A.~H. Altalhi, and F.~Herrera, ``Mixed opinion
  dynamics based on degroot model and hegselmann--krause model in social
  networks,'' \emph{IEEE Transactions on Systems, Man, and Cybernetics:
  Systems}, vol.~53, no.~1, pp. 296--308, 2022.

\bibitem{hazra2017using}
A.~Hazra, ``Using the confidence interval confidently,'' \emph{Journal of
  thoracic disease}, vol.~9, no.~10, p. 4125, 2017.

\bibitem{7073630}
R.~R. Yager, ``Golden rule and other representative values for atanassov type
  intuitionistic membership grades,'' \emph{IEEE Transactions on Fuzzy
  Systems}, vol.~23, no.~6, pp. 2260--2269, 2015.

\bibitem{7934318}
------, ``Multi-criteria decision making with interval criteria satisfactions
  using the golden rule representative value,'' \emph{IEEE Transactions on
  Fuzzy Systems}, vol.~26, no.~2, pp. 1023--1031, 2018.

\bibitem{1988Structuridentificatione}
M.~Sugeno and G.~T. Kang, ``Structure identification of fuzzy model,''
  \emph{Fuzzy Sets and Systems}, vol.~28, no.~1, pp. 15--33, 1988.

\bibitem{10480235}
Q.~Jia and E.~Herrera-Viedma, ``3-d golden rule representative value in
  decision-making systems,'' \emph{IEEE Transactions on Fuzzy Systems},
  vol.~32, no.~6, pp. 3729--3740, 2024.

\bibitem{2010Improved}
G.~Y. Bao, X.~L. Lian, M.~He, and L.~L. Wang, ``Improved two-tuple linguistic
  representation model based on new linguistic evaluation scale,''
  \emph{Control \& Decision}, vol.~25, no.~5, pp. 780--784, 2010.

\bibitem{wang2022research}
B.~Wang, S.~Li, J.~Mu, X.~Hao, W.~Zhu, and J.~Hu, ``Research advancements in
  key technologies for space-based situational awareness,'' \emph{Space:
  Science \& Technology}, 2022.

\end{thebibliography}
\begin{IEEEbiography}[{\includegraphics[width=1in,height=1.25in,clip,keepaspectratio]{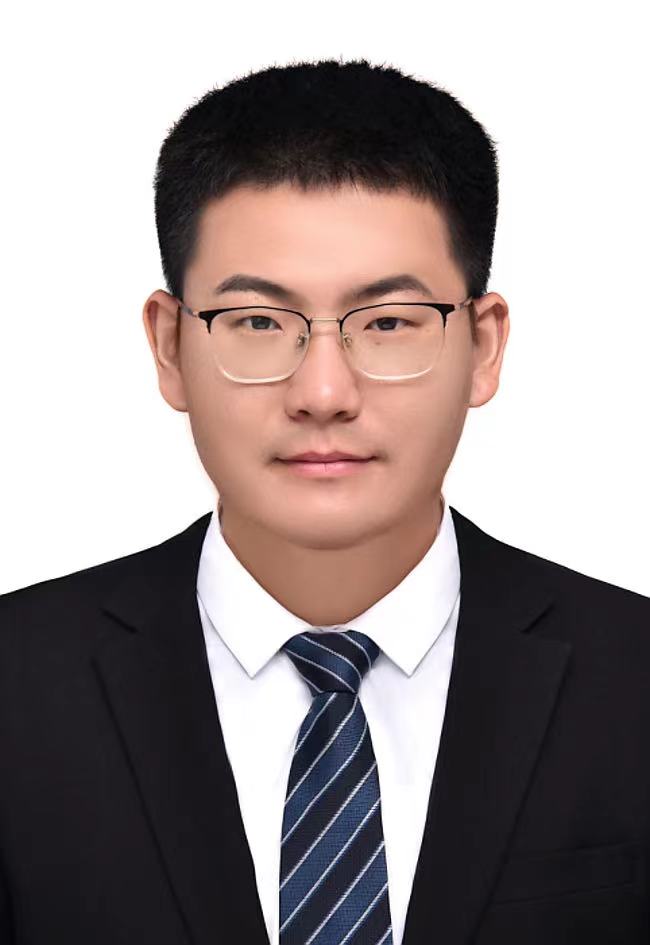}}]{Qianlei Jia} received the B.Eng. degree, M.Eng. degree, and Ph.D. degree from Northwestern Polytechnical University, Xi'an, China, in 2015, 2018, and 2022. He was a Research Associate with the School of
Mechanical and Aerospace Engineering, Nanyang Technological University (NTU), Singapore, in 2022, where he is currently a Postdoctoral.
\par His research interests include fuzzy theory, decision-making support systems, and information fusion.
\end{IEEEbiography}

\begin{IEEEbiography}[{\includegraphics[width=1in,height=1.25in,clip,keepaspectratio]{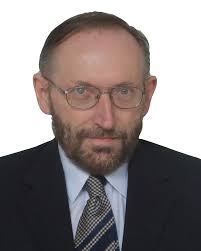}}]{Witold Pedrycz} (Life Fellow, IEEE) received the M.
Sc., Ph. D., and D. Sc. degrees from Silesian University of Technology, Gliwice, Poland, in 1977, 1980, and 1984, respectively. He is a Professor with the Department of Electrical and Computer Engineering, University of Alberta, Edmonton, AB, Canada. He is also with the Systems Research Institute, Polish
Academy of Sciences, Warsaw, Poland, where he
was elected as a foreign member in 2009. He also holds an appointment as Special Professor with the School of Computer Science, University of Nottingham, Nottingham, U.K. His current research interests include computational intelligence, fuzzy modeling and granular computing, knowledge discovery and data mining, fuzzy control, software engineering. He is also the author of 14 research monographs that cover various aspects of computational intelligence and software engineering.
\par He is the Editor-in-Chief of \emph{Information Sciences}. He is also an Associate Editor of \emph{IEEE Transactions on Fuzzy Systems}, \emph{IEEE Transactions on Systems,
Man, and Cybernetics: Systems}. He has edited a number of volumes. He has
been a member of numerous program committees of IEEE conferences in the
area of fuzzy sets and neurocomputing. He received the prestigious Norbert
Wiener Award from the IEEE Systems, Man, and Cybernetics Council in
2007. He received the IEEE Canada Computer Engineering Medal in 2008
and the Cajastur Prize for Soft Computing in 2009 from the European Centre
for Soft Computing for pioneering and multifaceted contributions to granular
computing. He is a Fellow of the Royal Society of Canada.
\end{IEEEbiography}

\end{document}